\documentclass{article}

\usepackage{microtype}
\usepackage{graphicx}
\usepackage{subfigure}
\usepackage{booktabs} 

\usepackage{hyperref}

\newcommand{\wukong}{Wukong\xspace}
\newcommand{\fulltitle}{\wukong: Towards a Scaling Law for Large-Scale Recommendation}
\newcommand{\neo}{Neo\xspace}
\newcommand{\sota}{state-of-the-art\xspace}

\newcommand{\gflop}{GFLOP/example\xspace}
\newcommand{\rll}{Relative LogLoss\xspace}
\newcommand{\nparams}{\#Params\xspace}

\newcommand\camera[1]{{\color{black} <#1>}}
\newcommand\yxedit[1]{{\color{black} #1}} 

\usepackage[accepted]{icml2024}

\usepackage{amsmath}
\usepackage{amssymb}
\usepackage{mathtools}
\usepackage{amsthm}

\usepackage[capitalize,noabbrev]{cleveref}

\theoremstyle{plain}

\theoremstyle{definition}

\theoremstyle{remark}

\usepackage[textsize=tiny]{todonotes}

\icmltitlerunning{\fulltitle}

\usepackage{xspace}
\usepackage{graphicx} 
\usepackage{xspace}
\usepackage{xcolor}
\usepackage{water-curtain-cave}

\begin{document}

\twocolumn[
\icmltitle{\fulltitle}



\icmlsetsymbol{equal}{*}

\begin{icmlauthorlist}
\icmlauthor{Buyun Zhang}{equal,meta}
\icmlauthor{Liang Luo}{equal,meta}
\icmlauthor{Yuxin Chen}{equal,meta} \\
\icmlauthor{Jade Nie}{meta}
\icmlauthor{Xi Liu}{meta}
\icmlauthor{Daifeng Guo}{meta} 
\icmlauthor{Yanli Zhao}{meta}
\icmlauthor{Shen Li}{meta} \\
\icmlauthor{Yuchen Hao}{meta}
\icmlauthor{Yantao Yao}{meta} 
\icmlauthor{Guna Lakshminarayanan}{meta} \\
\icmlauthor{Ellie Dingqiao Wen}{meta}
\icmlauthor{Jongsoo Park}{meta} \\
\icmlauthor{Maxim Naumov}{meta}
\icmlauthor{Wenlin Chen}{meta}
\end{icmlauthorlist}

\icmlaffiliation{meta}{Meta AI}

\icmlcorrespondingauthor{Buyun Zhang}{buyunz@meta.com}
\icmlcorrespondingauthor{Liang Luo}{liangluo@meta.com}
\icmlcorrespondingauthor{Yuxin Chen}{yuxinc@meta.com}

\icmlkeywords{Large scale recommendation system, Scaling law}

\vskip 0.3in
]



\printAffiliationsAndNotice{\icmlEqualContribution} 

\begin{abstract}
Scaling laws play an instrumental role in the sustainable improvement in model quality. Unfortunately, recommendation models to date do not exhibit such laws similar to those observed in the domain of large language models, due to the inefficiencies of their upscaling mechanisms. This limitation poses significant challenges in adapting these models to increasingly more complex real-world datasets. In this paper, we propose an effective network architecture based purely on stacked factorization machines, and a synergistic upscaling strategy, collectively dubbed \wukong, to establish a scaling law in the domain of recommendation. \wukong's unique design makes it possible to capture diverse, any-order of interactions simply through taller and wider layers.
We conducted extensive evaluations on six public datasets, and our results demonstrate that \wukong consistently outperforms \sota models quality-wise. Further, we assessed \wukong's scalability on an internal, large-scale dataset. The results show that \wukong retains its superiority in quality over \sota models, while holding the scaling law across two orders of magnitude in model complexity, extending beyond 100 \gflop, where prior arts fall short.

\end{abstract}

\begin{figure}[t]
    \centering
    \includegraphics[width=1\linewidth]{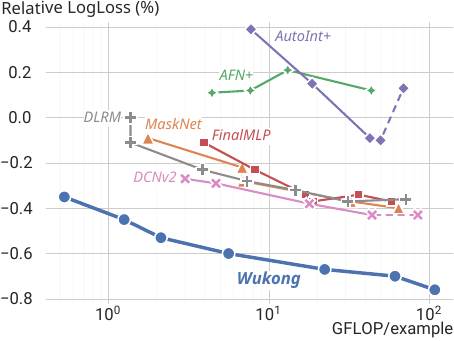}
    \caption{
    \wukong outperforms existing \sota models while demonstrating a scaling law in the recommendation domain across two orders of magnitude in model complexity, extending beyond 100 GFLOP/example.} 
    \label{fig:scaling}
\end{figure}


\section{Introduction}

Deep learning-based recommendation systems (DLRS) power a wide range of online services today~\cite{dlrm, wang2021dcn, persia, liu2022monolith, covington2016deep}. 

Modern DLRS are designed to process a blend of continuous dense features, such as date, and categorical sparse features, like user clicked posts history. Each sparse feature is transformed into a dense embedding representation through a trainable embedding lookup table. These dense embeddings are then fed into an  interaction component, designed to capture the intricate interactions between features. 

While existing models demonstrate promising accuracy on smaller datasets, 
their capability to adapt to the scale and intricacy of substantially larger datasets,
and to sustain continuous quality improvement as these models scale up, remains less certain.
This scalability is increasingly crucial, as modern datasets have seen exponential growth. 
For example, production datasets today might contain hundreds of billions of training examples~\cite{wang2021dcn}.
Furthermore, foundational models~\cite{bommasani2021opportunities} need to operate at scale to handle larger and multiple complex input sources at the same time.
Thus, the need for a DLRS that can both upscale and downscale effectively, adjusting to varying dataset sizes and computational constraints, is paramount. This scalability is encompassed in what is known as a "scaling law"~\cite{kaplan2020scaling}.



To date, the primary trend of DLRS up-scaling is through \textit{sparse scaling}, i.e., expanding the sizes of embedding tables (more rows and/or higher dimensions) for less collision and better expressiveness. Consequently, DLRS have reached trillions of parameters~\cite{kang2020learning, neo, persia} with embedding tables dominating the parameter count. Unfortunately, this traditional way of up-scaling has a few practical drawbacks. 
Merely expanding the sparse component of a model does not enhance its ability to capture the complex interactions among an increasing number of features.
Moreover, this trend notably diverges from the trend of hardware advancements, as most improvements in the next generation accelerators lie in the compute capacity~\cite{luo2018parameter, luo2017motivating}, which embedding table lookups cannot utilize. Thus, simply expanding embedding table leads to prohibitive infrastructure costs with suboptimal accelerator utilization, especially in distributed settings~\cite{dmt}. 

Our work aims to find an alternative scaling mechanism for recommendation models, that can establish a scaling law, similar to that established in the LLM domain. Namely, we would like to devise a unified architecture whose quality can be continuously improved in conjunction with dataset size, compute and parameter budgets, with a synergistic strategy.

We focus on upscaling interaction components, dubbed \textit{dense scaling}, to mitigate the quality and efficiency drawbacks from sparse scaling. 
However, existing models cannot benefit from this paradigm for various reasons.
For example, DLRM lacks the ability to capture \camera{higher-order} interactions; DCNv2 and AutoInt+ lack strategy for effective upscaling, leading to rapidly diminishing returns when scaling up; further, 
even with modern tricks like residual connection~\cite{resnet}, layernorm~\cite{ba2016layer}, gradient clip~\cite{pascanu2013difficulty}), up-scaling existing models is prone to training stability issues~\cite{tang2023clippy}.
To establish a scaling law for recommendation models, we propose \wukong, a simple interaction architecture that exhibits effective dense scaling properties.
Inspired by the principles of binary exponentiation, 
our key innovation is to use a series of stacked Factorization Machines (FMs) to efficiently and scalably capture any-order feature interactions.
In our design, each FM is responsible of capturing second order interactions with respect to its inputs, and the outputs from these FMs are subsequently transformed by MLPs into new embeddings, which encode the interactions results and serve as inputs to the next layers. 



We evaluated Wukong's performance using six public datasets and a large-scale internal dataset. 
The results demonstrate that \wukong outperforms \sota models across all public datasets in terms of AUC,
indicating the effectiveness of \wukong's architecture and its ability to generalize across a wide range of recommendation tasks and datasets.
In our internal dataset, \wukong not only significantly outperforms existing models in terms of quality at comparable levels of complexity but also shows continuous enhancements in quality when scaled up across two orders of magnitude in model complexity, extending beyond 100 GFLOP/example, where prior arts fall short.  





\section{Related Work}

\para{Deep Learning Recommendation Systems (DLRS)}
Existing DLRS share a similar structure. A typical model consists of a sparse and a dense component.
The sparse component is essentially embedding lookup tables that transform sparse categorical features into dense embeddings, whereas the dense component is responsible for capturing interactions among these embeddings to generate a prediction.

\para{Dense Interaction Architectures} 
Capturing interaction between features is the key to DLRS effectiveness, and we highlight some of the prior arts. 
AFN+ \cite{afn} transforms features into a logarithmic space to capture arbitrary order of interactions;
AutoInt+ \cite{song2019autoint} uses multi-head self-attention; 
DLRM and DeepFM \cite{dlrm,guo2017deepfm} leverage Factorization Machines (FM) \cite{fm} to explicitly capture second order interactions;
HOFM~\cite{blondel2016hofm} optimizes FM to efficiently capture higher order of interactions;
DCNv2 \cite{wang2021dcn} uses CrossNet, which captures interactions via stacked feature crossing, which can be viewed as a form of elementwise input attention.
FinalMLP \cite{mao2023finalmlp} employs a bilinear fusion to aggregate results from two MLP streams, each takes stream-specific gated features as input.
MaskNet \cite{wang2021masknet} adopts a series of MaskBlocks for interaction capture, applying ``input attention'' to the input itself and intermediate activations of DNN;
xDeepFM \cite{lian2018xdeepfm} combines a DNN with a Compressed Interaction Network, which captures interactions through outer products and compressing the results with element-wise summation.  





\para{Scaling up DLRS}
~\cite{kang2020learning, neo, persia} provides mechanisms on sparse scaling. ~\cite{shin2023scaling} focuses on scaling up user representation models, with the largest model reported having a total compute less than 0.1 PF-days, ~\cite{zhang2023scaling} aims to improve sequence modeling on the user side, with the largest model reported having less than 0.8B parameters. Additionally ~\cite{ardalani2022understanding} studied the scaling law of DLRM, which is incorporated as a baseline in our work and further scaled up in our experiments. Orthogonally, \cite{zhao2023breaking} proposes a user-centric ranking formulation to improve scalability; ~\cite{guo2023embedding} provided insights on sparse scaling, demonstrating limits on prior arts and is complement to our work. Further, VIP5~\cite{geng2023vip5} leverages existing scaling laws in LLMs to apply a multimodal LLM to recommendation, however, ~\cite{lin2023can} points out that further study is needed to verify whether larger implies better in LLM-powered recommenders, while ~\cite{huang2024foundation} suggests evaluations on more diverse datasets are needed to for a conclusion.





\section{Design of \wukong}


We keep two objectives in mind when designing \wukong's architecture: (1) to effectively capture the intricate high-order feature interactions; and (2) to ensure \wukong's quality scale gracefully with respect to dataset size, \gflop and parameter budgets.

\subsection{Overview}
In \wukong, categorical and dense features initially pass through an \textbf{Embedding Layer} (Sec. \ref{design:embedding}), which transforms these inputs into {\em Dense Embeddings}. 

\begin{figure}[t]
    \centering
    \includegraphics[width=.8666\linewidth]{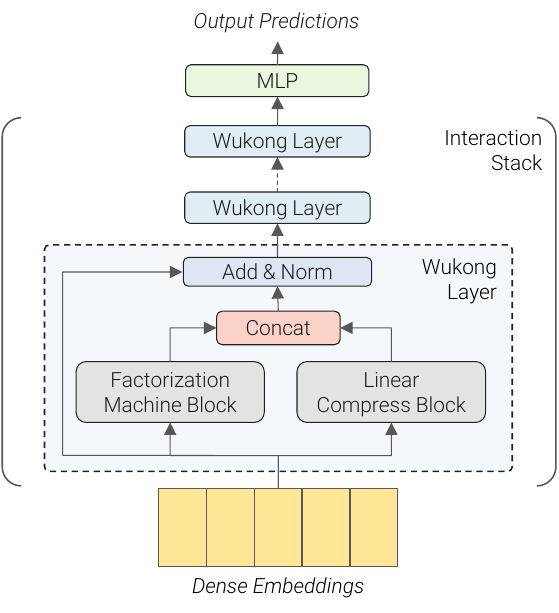}
    \caption{Wukong employs an interaction stack to capture feature interactions. 
    Each layer in the stack consists of a Factorization Machine Block and a Linear Compress Block.
    }
    \label{fig:teaser}
\end{figure}

As shown in Figure~\ref{fig:teaser}, \wukong subsequently adopts an \textbf{Interaction Stack} (Sec. \ref{design:stack}), a stack of unified neural network layers to capture the interaction between embeddings. 
The Interaction Stack draws inspiration from the concept of binary exponentiation, allowing each successive layer to capture exponentially higher-order interactions.
Each layer in the Interaction Stack consists of a \textbf{Factorization Machine Block} (FMB, Sec. \ref{design:fmb}) and a \textbf{Linear Compression Block} (LCB, Sec. \ref{design:lcb}). FMB and LCB independently take in input from last layer and their outputs are ensembled as the output for the current layer. 
Following the interaction stack is a final Multilayer Perceptron (MLP) layer that maps the interaction results into a prediction.

\subsection{Embedding Layer}
\label{design:embedding}

Given a multi-hot categorical input, an embedding table maps it to a dense embedding.
This process involves a series of lookups, each corresponding to a ``hot'' dimensions within the input.
The lookup results are then aggregated using a pooling operation (usually a summation).


In our design, the embedding dimension is standardized 
for all embeddings generated by the Embedding Layer, known as the global embedding dimension $d$. 
To accommodate the varying significance of different features, multiple embeddings are generated for each feature that is deemed significant.
Less important features are allocated smaller underlying embedding dimensions. 
These smaller embeddings are then collectively grouped, concatenated, and transformed into $d$-dimensional embeddings using a MLP. 

Dense inputs are transformed by an MLP into latent embeddings that share the same $d$ dimension, and are joined with the embedding outputs of categorical input. This yields an output tensor of size $X_0 \in \mathbb{R}^{n \times d}$, where $n$ is the total number of embeddings from the dense and sparse part. $X_0$ is then ready to be further processed by the Interaction Stack.

Note that unlike conventional approaches like DCN \cite{wang2021dcn}, we interpret each embedding vector as a whole unit (detailed later), and hence our representation of $X_0 \in \mathbb{R}^{n \times d}$ as opposed to $X_0 \in \mathbb{R}^{nd}$.

\subsection{Interaction Stack}
\label{design:stack}
The interaction modules stack $l$ identical interaction layers, where each layer captures progressively higher-order feature interactions using Factorization Machines (FMs).


An interaction layer has two blocks in parallel: a Factorization Machine Block (FMB) and a Linear Compression Block (LCB). 
FMB computes feature interactions between input embeddings of the layer, 
and LCB simply forwards linearly compressed input embeddings of the layer. 
The outputs of FMB and LCB are then concatenated. 

For layer $i$ in the stack, its results can contain feature interactions with arbitrary order from 1 to $2^{i}$. This can be simply shown by induction.
Let's assume the input of layer $i$ contains interactions of order from 1 to $2^{i-1}$, which is true for the first layer (i.e. $i=1$). 
Since FMB generates $(o_1 + o_2)$-order feature interactions given $o1$ and $o2$-order interactions, then we have immediately the output of layer $i$ containing 1 to $2^{i}$-order interactions, with the lower bound achieved from the output of LCB and the upper bound achieved by the FM interacting two $2^{i-1}$-order interactions from the input.


To help stabilize training, we also adopt residual connections across layers, followed by layer normalization (LN). Putting everything together, we have
\begin{equation*}
\label{eq:interact-stack}
X_{i+1} = \mathrm{LN(concat(FMB_i}(X_i), \mathrm{LCB_i}(X_i)) + X_i)
\end{equation*}


Depending on the specific configurations of FMB and LCB, 
$X_i$ may have a different number of embeddings than $X_{i+1}$, 
which usually happens at the first layer.
To handle this case, the residual can be linearly compressed to match the shape.

\subsection{Factorization Machine Block (FMB)}
\label{design:fmb}
A FMB contains a FM followed by a MLP. The FM is used to capture explicit feature interactions of the input embeddings, with the output being a 2D interaction matrix where each element represents the interaction between a pair of embeddings.
This interaction matrix is flattened and converted to a vector with shape of $(n_{F} \times d)$ through the MLP, and reshaped to $n_F$ embeddings for later use.

Operationally, a FMB does the following:
\begin{equation*}
\mathrm{FMB}(X_i) = \mathrm{reshape(MLP(LN(flatten(FM}(X_i)))))
\end{equation*}

\wukong's FM module is fully customizable: for example, in the most basic version, 
we followed the FM design in \cite{dlrm}, i.e.,
taking the dot product between all pairs of embedding vectors, $FM(X) = XX^T$.
We discuss more optimized FM designs in Sec. \ref{design:optimized-fm}.


\subsection{Linear Compress Block (LCB)}
\label{design:lcb}
LCB simply linearly recombines embeddings without increasing interaction orders, which is critical in ensuring that the invariance of interaction order is maintained throughout the layers. Specifically, it guarantees that the $i$-th interaction layer captures interaction orders ranging from 1 to $2^i$. 
The operation performed by a LCB can be described as follows: 

\begin{equation*}
\mathrm{LCB}(X_i) = W_{L}X_i
\end{equation*}

where $W_{L} \in \mathbb{R}^{n_{L} \times n_i}$ is a weight matrix,
$n_{L}$ is a hyperparameter indicating the number of compressed embeddings, and
$n_i$ is the number of input embeddings of layer $i$.




\subsection{Optimized FM}
\label{design:optimized-fm}
FM's computation and storage complexity grows quadratically with the number of embeddings with the pair-wise dot product, and This quickly becomes prohibitive on real-world datasets with thousands of features. 

To allow effective feature interaction while lowering compute cost, we adopt a similar scheme to ~\cite{FeatureF20:online, dcpp} that leverage low-rank property in pair-wise dot product matrix, which was observed in many real-world datasets~\cite{wang2021dcn}.


When $d <= n$, the dot-product interaction $XX^T$ is a $d$-rank matrix, which is often the case on large datasets whose number of features is larger than the embedding dimension. 
Therefore, we can effectively reduce the size of output matrix from $n \times n$ to $n \times k$, where $k$ is a hyperparameter, by multiplying $XX^T$ with a learnable projection matrix $Y$ of shape $n \times k$ (i.e., computing $XX^TY$) without loss of information in theory. This reduces memory requirement to store the interaction matrix.
We can then take advantage of the associative law to compute $X^TY$ first, further reducing compute complexity from $O(n^2d)$ to $O(nkd)$ with $k << n$.

Furthermore, to enhance the model quality, the projection matrix $Y$ can be made attentive to the input by processing linearly compressed input through a MLP.
We use the optimized FM in our following experiments by default,
unless mentioned otherwise. 

\subsection{Complexity Analysis}
\label{design:complexity}
We assume each layer in the Interaction Stack uses the same hyperparameters, and the largest FC in the MLP has size $h$.

For the first layer,
the time complexity of FMB is the sum of the FM and the MLP,
which is $O(nkd) \approx O(ndh)$ and $O(nkh + h^2 + n_Fdh) \approx O(ndh + h^2)$, respectively.
The time complexity of LCB is $O(nn_Ld) \approx O(ndh)$.
For subsequent layers,
the time complexity is $O(n'dh + h^2)$, where $n' = n_L + n_F$.
Hence, the total time complexity of \wukong is $O(ndh + ln'dh + h^2) \approx O(ndhlogn+h^2)$.

\subsection{Scaling \wukong}
\label{design:scaling}
We now summarize the main hyperparameters that are related to scale up and later we describe our efforts to upscaling \wukong with respect to these hyperparameters. 

\begin{packed_itemize}
\item $l$: number of layers in the Interaction Stack.
\item $n_F$: number of embeddings generated by FMB 
\item $n_L$: number of embeddings generated by LCB 
\item $k$: number of compressed embeddings in optimized FM
\item $MLP$: number of layers and FC size in the MLP of FMB
\end{packed_itemize}

During scaling up, we initially focus on increasing $l$ to enable the model to capture higher-order interactions.
Following this, we enlarge other hyperparameters to augment the model's capacity of capturing broader range of interactions.

\subsection{Intuition Behind \wukong's Enhanced Effectiveness}



Compared to existing work using FM as their primary interaction architecture, 
\wukong's innovative approach of stacking FMs greatly enhances the conventional FM's capability.
This allows \wukong to capture interactions of any order, making it highly effective for large-scale, complex datasets that require higher-order reasoning.
While there are efforts towards high-order FM, Wukong's exponential rate of capturing high-order interactions offers great efficiency, bypassing the linear complexity seen in HOFM and avoiding the costly outer product in xDeepInt.

While MLPs have shown limitations in implicitly capturing interactions \cite{beutel2018latent}, Wukong diverges from approaches that rely on MLPs for interaction capture. Instead, Wukong primarily employs MLPs to transform the results of interactions into embedding representations, which are then used for further interactions. This distinct use of MLPs enhances the model's ability to process and interpret complex, heterogeneous features effectively.


Additionally, \wukong treats each embedding as a single unit, focusing on embedding-wise interactions. This approach significantly reduces computational demands compared to architectures that capture element-wise interactions. 

\section{Implementation}
\label{sec:imple}


This section discusses practices to effectively train high-complexity \wukong on large-scale datasets. 

Overall, distributed training is required to make \wukong training feasible. For the embedding layer, we use a column-wise sharded embedding bag implementation provided by \neo~\cite{neo} and NeuroShard~\cite{neuroshard}. On the dense part, we balance the trade-off between performance and memory capacity by adopting FSDP~\cite{fsdp} and tune the sharding factor so that the model fits in the memory without creating too much redundancy.

To enhance training efficiency,
we employ automatic operator fusion through to improve training performance. 
In addition, we aggressively apply quantization to reduce compute, memory, and communication overheads simultaneously. Specifically, we train \wukong's embedding tables in FP16, and communicate embedding lookup results in FP16 in the forward pass and BF16 in the backward pass; we use BF16 quantization during the transport gradients for dense parameters in the backward pass. 

\section{Overview of Evaluations}
We evaluate \wukong using six public datasets and an internal dataset,
details of which are summarized in Table~\ref{tab:datasets}. 
The results of these evaluations are organized in two sections.

In Section~\ref{sec:eval-public}, we evaluate on six public datasets, 
focusing on demonstrating the effectiveness of \wukong in the low complexity realm.
Our results show that \textbf{\wukong surpasses previous \sota methods across all six datasets, demonstrating its effectiveness.}

In Section~\ref{sec:eval-prod}, we evaluate on our large-scale in-house dataset to demonstrate the scalability of \wukong. The dataset contains 30 times more samples and 20 times more features compared to one of the largest dataset Criteo. 
Our results reveals that \textbf{(1) \wukong consistently outperforms all baseline models in terms of both model quality and runtime speed, maintaining this superiority across all complexity scales; (2) \wukong exhibits a better scaling trend in comparison to baseline models.}
We also conduct an ablation study to gain understanding of the individual contributions and the effectiveness of each component within \wukong.

\begin{table}[t]
\centering
\footnotesize\fontfamily{\robotofamily}\selectfont
\setlength{\tabcolsep}{1.5666em}
\begin{tabular}{lrr}
\toprule
     & \textbf{\#Samples}     & \textbf{\#Features}        \\
\midrule

Frappe & 0.29M & 10 \\
MicroVideo & 1.7M & 7 \\
MovieLens Latest & 2M & 3 \\
KuaiVideo & 13M & 8 \\
TaobaoAds & 26M & 21 \\  
Criteo Terabyte & 4B &  39 \\

\addlinespace[2px]
\hdashline
\addlinespace[4px]

Internal & 146B & 720 \\

\bottomrule
\end{tabular}
\caption{
Statistics of our evaluation datasets.
}
\label{tab:datasets}
\end{table}

\section{Evaluation on Public Datasets}
\label{sec:eval-public}
\begin{table*}[t]

\footnotesize\fontfamily{\robotofamily}\selectfont
\setlength{\tabcolsep}{.5616666em}

\begin{tabular}{lc@{\hskip -.0266em}cc@{\hskip -.0266em}cc@{\hskip -.0266em}cc@{\hskip -.0266em}cc@{\hskip -.0266em}cc@{\hskip -.0266em}c}
\toprule
         & \multicolumn{2}{c}{\textbf{Frappe}} & \multicolumn{2}{c}{\textbf{MicroVideo}} & \multicolumn{2}{c}{\textbf{MovieLens L.}} & \multicolumn{2}{c}{\textbf{KuaiVideo}} & \multicolumn{2}{c}{\textbf{TaobaoAds}} & \multicolumn{2}{c}{\textbf{Criteo TB}} \\
         & AUC                           & {\color[HTML]{656565} LogLoss}                       & AUC                  & {\color[HTML]{656565} LogLoss}              & AUC                           & {\color[HTML]{656565} LogLoss}              & AUC                  & {\color[HTML]{656565} LogLoss}              & AUC                  & {\color[HTML]{656565} LogLoss}              & AUC                           & \multicolumn{1}{c}{{\color[HTML]{656565} LogLoss}} \\
\midrule

\addlinespace[2px]
\multicolumn{11}{l}{\textit{Baselines}} \\
\addlinespace[2px]

AFN+     & 0.9812          & {\color[HTML]{656565} 0.2340}       & 0.7220          & {\color[HTML]{656565} \underline{0.4142}} & 0.9648          & {\color[HTML]{656565} 0.3109}       & 0.7348          & {\color[HTML]{656565} 0.4372}       & 0.6416          & {\color[HTML]{656565} 0.1929}       & 0.8023                        & {\color[HTML]{656565} 0.1242}       \\
AutoInt+ & 0.9806          & {\color[HTML]{656565} 0.1754}       & 0.7155          & {\color[HTML]{656565} 0.4203}       & 0.9693          & {\color[HTML]{656565} 0.2178}       & 0.7297          & {\color[HTML]{656565} 0.4376}       & 0.6437          & {\color[HTML]{656565} 0.1930}       & 0.8073                        & {\color[HTML]{656565} 0.1233}       \\
DCNv2    & 0.9774          & {\color[HTML]{656565} 0.2325}       & 0.7187          & {\color[HTML]{656565} 0.4162}       & 0.9683          & {\color[HTML]{656565} 0.2169}       & 0.7360          & {\color[HTML]{656565} 0.4383}       & 0.6457          & {\color[HTML]{656565} \underline{0.1926}} & 0.8096                        & {\color[HTML]{656565} 0.1227}       \\
DLRM     & 0.9846          & {\color[HTML]{656565} 0.1465}       & 0.7173          & {\color[HTML]{656565} 0.4179}       & 0.9685          & {\color[HTML]{656565} 0.2160}       & 0.7357          & {\color[HTML]{656565} 0.4382}       & 0.6430          & {\color[HTML]{656565} 0.1931}       & 0.8076                        & {\color[HTML]{656565} 0.1232}       \\
FinalMLP & \textbf{0.9868} & {\color[HTML]{656565} \underline{0.1280}} & 0.7247          & {\color[HTML]{656565} 0.4147}       & \textbf{0.9723} & {\color[HTML]{656565} 0.2211}       & 0.7374          & {\color[HTML]{656565} 0.4435}       & 0.6434          & {\color[HTML]{656565} 0.1928}       & 0.8096                        & {\color[HTML]{656565} 0.1226}       \\
MaskNet  & 0.9816          & {\color[HTML]{656565} 0.1701}       & 0.7255          & {\color[HTML]{656565} 0.4157}       & 0.9676          & {\color[HTML]{656565} 0.2383}       & 0.7376          & {\color[HTML]{656565} 0.4372}       & 0.6433          & {\color[HTML]{656565} 0.1927}       & 0.8100                        & {\color[HTML]{656565} 0.1227}       \\
xDeepFM  & 0.9780          & {\color[HTML]{656565} 0.2441}       & 0.7167          & {\color[HTML]{656565} 0.4172}       & 0.9667          & {\color[HTML]{656565} 0.2089}       & 0.7118          & {\color[HTML]{656565} 0.4565}       & 0.6342          & {\color[HTML]{656565} 0.1961}       & 0.8084          & {\color[HTML]{656565} 0.1229}       \\

\addlinespace[2px]
\hdashline
\addlinespace[4px]

\multicolumn{11}{l}{\textit{Ours}} \\
\addlinespace[2px]

Wukong   & \textbf{0.9868} & {\color[HTML]{656565} 0.1757}       & \textbf{0.7292} & {\color[HTML]{656565} 0.4148}       & \textbf{0.9723} & {\color[HTML]{656565} \underline{0.1794}} & \textbf{0.7414} & {\color[HTML]{656565} \underline{0.4367}} & \textbf{0.6488} & {\color[HTML]{656565} 0.1954}       & \textbf{0.8106}               & {\color[HTML]{656565} \underline{0.1225}}
\\

\bottomrule
\end{tabular}
\caption{
Evaluation results on six public datasets. The model with \textbf{best AUC} and {\color[HTML]{656565} \underline{best LogLoss}} on each dataset are highlighted.
}
\label{tab:eval-public}

\end{table*}

In this section, we aim to demonstrate the effectiveness of \wukong across a variety of public datasets. Unless noted otherwise, we use the preproc provided by the BARS benchmark~\cite{barsbench} for consistency with prior work.

\subsection{General Evaluation Setup}
\subsubsection{Datasets}
\textbf{Frappe \cite{frappe}} is an app usage log. This datasets predicts whether a user uses the app with the given contexts.

\textbf{MicroVideo \cite{microvideo}} is a content understanding-based dataset provided by THACIL work containing interactions between users and micro-videos. This log contains multimodal embeddings,
together with traditional features.

\textbf{MovieLens Latest \cite{movielens}} is a well known dataset that contains users' ratings on movies. 

\para{KuaiVideo \cite{kuaivideo}} is the competition dataset released by Kuaishou. The dataset is used to predict the click probability of a user on new micro-videos. This dataset also contains content understanding-based embeddings along with other categorical and float features.

\para{TaobaoAds \cite{taobaoads}}
This dataset includes 8 days of ads click through rate (CTR) prediction on Taobao.

\para{Criteo Terabyte \cite{criteo}}
This dataset contains 24 days of ads click feedback.
We used the last day of data for testing.

\subsubsection{Baselines}
\label{eval:baselines}
We benchmark \wukong against seven widely recognized \sota models used in both academia and industry, including
AFN+ \cite{afn}, AutoInt+ \cite{song2019autoint}, DLRM \cite{dlrm}, DCNv2 \cite{wang2021dcn},
FinalMLP \cite{mao2023finalmlp}, MaskNet \cite{wang2021masknet}
and xDeepFM \cite{lian2018xdeepfm}.

\subsubsection{Metrics}

\para{AUC} Area Under the Curve (AUC) measures the model's ability to correctly classify positives and negatives across all thresholds. Higher the better. We use AUC as the basis for hyperparameter tuning and topline metric for reporting, following recommendation conventions~\cite{DisplayA24:online, blondel2016hofm, song2019autoint, wang2021dcn, barsbench, mao2023finalmlp}.

\para{LogLoss} The log loss quantifies the penalty based on how far the prediction is from the actual label. Lower the better.

\subsection{Model-Specific Setup}
For the five smaller datasets, aside from Criteo,
we adopted the public BARS evaluation framework \cite{bars,bars2}.
We directly use the best searched model configs on BARS whenever possible,
and use the provided model default hyperparameters for the rest.
In addition to the default embedding dimension provided in the framework,
we further test an embedding dimension of 128 and report whichever of these two configurations yielded better results.
For \wukong, we tune the dropout rate and optimizer settings and compression of LCB to adapt to the number of features.

We leverage the larger Criteo dataset to evaluate the model performance on realistic online recommendation systems,
where one-pass training is performed.
In light of the new training setup, 
we conducted extensive grid search using the system described in Sec. \ref{sec:imple} for all baselines and \wukong to facilitate fair comparisons.
This exhaustive process involved nearly 3000 individual runs.
We provide the model-specific search space in Appendix \ref{appendix:criteo-search}.
The best searched model hyperparameters were later used as the base config in Sec~\ref{sec:eval-prod}.

\subsection{Results}

We summarize the results in Table~\ref{tab:eval-public}. Overall, \wukong is able to achieve state-of-the-art results in terms of AUC across all public datasets. 
This result demonstrates the effectiveness of \wukong's architecture and its ability to comprehend diverse datasets and to generalize across a wide range of recommendation tasks.




\section{Evaluation on an Internal Dataset}
\label{sec:eval-prod}
In this section, we show the scalability of \wukong and gain a deep understanding of how different individual components of \wukong contribute to its effectiveness, using a large-scale dataset which  enables the study for merging properties that is not seen in small, public datasets.


\subsection{Evaluation Setup}

\subsubsection{Dataset}
This dataset contains 146B entries in total and has 720 distinct features. Each feature describes a property of either the item or the user. There are two tasks associated with this dataset: {\em (Task1)} predicting whether a user has showed interested in an item (e.g., clicked) and {\em (Task2)} whether a conversion happened (e.g., liked, followed). 

\subsubsection{Metrics}

\para{\gflop} Giga Floating Point Operations per example (\gflop) quantifies the computational complexity during model training. 

\para{PF-days} The total amount of training compute equivalent to running a machine operating at 1 PetaFLOP/s for 1 day.

\para{\nparams} Model size measured by the number of parameters in the model. The sparse embedding table size was fixed to 627B parameters. 


\para{\rll} LogLoss improvement relative to a fixed baseline. 
We opt to use the DLRM with the basic config as the baseline. 
A 0.02\% \rll improvement is considered as significant on this dataset.
We report relative LogLoss on the last 1B-window during online training. 

\subsubsection{Baselines}
We adhere to the same baseline setup as detailed in Sec. \ref{eval:baselines}.
However, xDeepFM was not included in the reported results, 
due to the incompatibility of its expensive outer product operation with the large-scale dataset,
consistently causing out-of-memory issues even in minimal setups. 

\subsubsection{Training}

We used the best optimizer configuration found in our pilot study across all experiments,
i.e., Adam with lr=0.04 with beta1=0.9, beta2=1 for dense part and Rowwise Adagrad with lr=0.04 for sparse embedding tables. 
Models were trained and evaluated in an online training manner. We fix the embedding dimension to 160 across all runs.

We set the hyperparameters with the best configuration found on the Criteo Terabyte evaluation described in Sec. \ref{sec:eval-public} as a starting point, and gradually scale up parameter count for each model. We use a global batch size of 262,144 for all experiments. Each experiment was run on 128 or 256 H100 GPUs depending on the model size.




\subsection{Results}
We observed comparable results for both tasks, and report results for {\em Task1} in the main text, while the detailed results of {\em Task2} are provided in Appendix~\ref{appendix:scaleup}.

\para{Quality vs. Compute Complexity} In Fig. \ref{fig:scaling}, we depict the relationship between quality and compute complexity (empirically, $y = -100 + 99.56x^{0.00071}$). 
The results show that Wukong consistently outperforms all baselines across various complexity levels, achieving over 0.2\% improvement in LogLoss.
Notably, \wukong holds its scaling law across two orders of magnitude in model complexity -- 
approximately translating to a 0.1\% \yxedit{improvement} for every quadrupling of complexity.
Among baselines, AFN+, DLRM and FinalMLP tend to reach a plateau after a certain complexity level,
\yxedit{while AutoInt+, DCNv2 and MaskNet failed to further enhance quality \footnote{\yxedit{AutoInt+ and DCNv2 consistently faced significant training instability issue when further scaled up. 
AutoInt+ recovered from loss explosion, albeit with reduced model quality; while 
DCNv2 failed to recover, and its quality was estimated from performance before the explosion.
MaskNet was hindered by excessive memory consumption, leading to out-of-memory errors, blocking further scaling up.}}.}
Nonetheless, even DCNv2, the top-performing baseline, demands a 40-fold increase in complexity to match \wukong's quality.

\begin{figure}[t]
    \centering
    \includegraphics[width=1\linewidth]{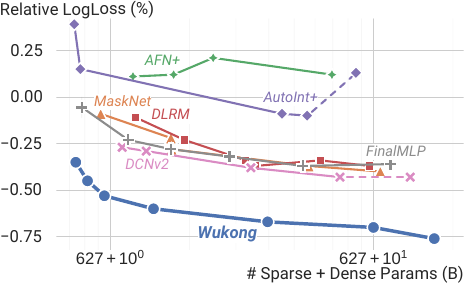}
    \caption{
    Scalability of \wukong with respect to \# parameters on the internal dataset. 
    }
    \label{fig:scaling-params}
\end{figure}

\para{Quality vs. Model Size} In Fig. \ref{fig:scaling-params}, we illustrate the correlation between model quality and model size.
Echoing the trends observed in compute complexity scaling above, 
\wukong consistently outperforms all baselines by roughly 0.2\% across all scales of model size.
while demonstrating a steady improvement trend up to over 637 billion parameters\footnote{We verified \wukong's effectiveness in both online and offline settings, and for brevity, we focus on reporting offline metrics.}

\para{Quality vs. Data Size} See Appendix~\ref{sec:qualityvsdatasetsize}.

\para{Model-Specifc Scaling} Throughout the scaling process, we employed distinct strategies per model. Detailed hyperparameter settings for each run are provided in Appendix~\ref{appendix:scaleup}. Scaling processes of each model are summarized as follows:

\parab{Wukong} We scaled up \wukong by tuning the hyperparameters detailed in Sec. \ref{design:scaling}.

\parab{AFN+} We scaled up AFN's hidden layers, ensemble DNN, and the number of logarithmic neurons. 
The results show that scaling up AFN does not improve model quality.

\parab{AutoInt+} We scaled up multi-head attention and the ensemble DNN.
Model quality of this model is initially worse than others, but improves notably when scaling up.

\parab{DLRM} We scaled up the top MLP. The results show that the quality starts saturated beyond 31 \gflop.

\parab{DCNv2} We scaled up both Cross Network and Deep Network.
Scaling up Cross Network did not yield any quality improvement.
The training stability of DCNv2 is worse than other models and we applied strict gradient clipping.

\parab{FinalMLP} We scaled up the two MLP streams and the Feature Selection modules.
The results show that the model quality improves in the low complexity region, but starts to saturate beyond 36 \gflop.

\parab{MaskNet} We tested both Parallel and Serial MaskNet, and found that the Parallel variant is better.
We decreased the initial reduction ratio to ensure the model has a runnable size,
 and progressively scaled up number of MaskBlocks, the DNN and the reduction ratio.


\subsection{Ablation}
\label{eval:abalation}






\para{Significance of Individual Components}
Our goal is to demonstrate the importance of FMB, LCB and the residual connection in \wukong's Interaction Stack.
To this end, we performed experiments in which each component was individually deactivated by zeroing out its results.

\begin{figure}[h]
    \centering
    \includegraphics[width=1\linewidth]{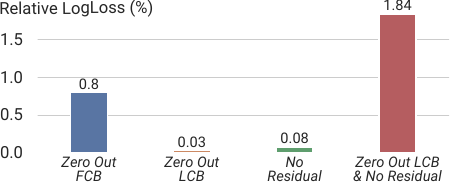}
    \caption{
    Significance of individual components.
    }
    \label{fig:ablation-component}
\end{figure}

As shown in Fig. \ref{fig:ablation-component},
nullifying FMB results in a large quality degradation.
Interestingly, the deactivation of either LCB or the residual leads to only a modest decline in quality, while disabling both causes a substantial degradation.
This observation implies that by zero-padding FMB outputs and incorporating a residual connection,
LCB can be simplified.

\begin{figure}[h]
    \centering
    \includegraphics[width=1\linewidth]{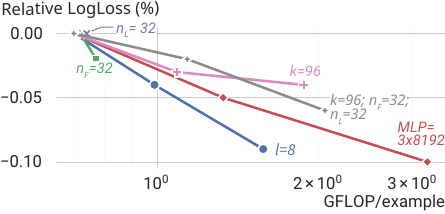}
    \caption{
    Impact of scaling individual components.
    }
    \label{fig:ablation-scaling}
\end{figure}

\para{Impact of Scaling Individual Components}
We aim to dissect the contributions in model quality when scaling up each hyperparameter within \wukong.
We started from a base configuration and proceeded to incrementally double each hyperparameter.
The results are depicted in Fig. \ref{fig:ablation-scaling}.
We observed that increasing the number of \wukong layers $l$ leads to a substantial uplift in model quality,
due to higher-order interactions being captured.
Additionally, augmenting the MLP size results in considerable performance enhancements.
Elevating $k$ and $n_F$ proves beneficial, while $n_L$ has plateaued for the base configuration.
Notably, a combined scale-up of $k, n_F, n_L$ delivers more pronounced quality improvements than scaling each individually.

\section{Discussions}
\para{Practically Serving Scaled-up Models}
Scaling up to high complexity presents notable challenges for real-time serving.
Potential solutions include training a multi-task foundation model to amortize costs: distilling knowledge from the large models into small, efficient ones for serving.



\para{Limitation and Future Work}
We also note limitations and caveats to our work, which can be goals in future work.

Understanding the exact limit of \wukong's scalability is an important area of research. Due to the massive compute requirement, we have not been able to reach a level of complexity where the limit applies.

While \wukong demonstrates superior quality in various evaluations, a comprehensive theoretical understanding of its underlying principles, particularly in contrast to architectures like transformers which share stacked dot product structure, remains an area that needs further exploration.

Additionally, \wukong's generalizability beyond recommendation, particularly in domains that involve heterogeneous input data sources similar to distinct features in recommendation, remains to be further explored and understood.

\section{Conclusion}
We proposed an effective network architecture, named Wukong.
We demonstrated that \wukong establishes a scaling law in the domain of recommendation that is not previously observed -- Wukong is able to efficiently scale up and down across two order of magnitude in compute complexity while maintaining a competitive edge over other state of the art models, making it a scalable architecture that can serve as a backbone from small vertical models to large foundational models across a wide range of tasks and datasets.

\section*{Impact Statement}
This paper presents work whose goal is to advance the field of Machine Learning. There are many potential societal consequences of our work, none which we feel must be specifically highlighted here.


\bibliographystyle{icml2024}
\bibliography{references}

\appendix
\section{Model-Specific Grid Search Space on Criteo}
\label{appendix:criteo-search}
We use Adam for dense arch optimization and use Rowwise AdaGrad for sparse arch optimization with a linear warmup period for the first 10\% steps.
We use $8 * 16384 = 131,072$ global batch size. All models use ReLU for activation.
We opted to use 128 as embedding dimension, as it shows better results on all models in our pilot experiments.
We use FP32 in all runs. Due to the dataset volume and model size, we use~\cite{neo} as the sparse distributed training framework and data parallel for dense synchronization.

To facilitate fair comparisons,  we conducted extensive grid search (>3000 runs) over both general hyper-parameters and model-specific configs on Criteo Dataset. 

For all the models, both sparse and dense learning rate was separately tuned in $\{1e^{-3}, 1e^{-2}, 1e^{-1}\}$. For MLPs in all the models, the number of hidden layers ranged in $\{1, 2, 3, 4\}$ with their layer sizes in $\{512, 1024, 2048\}$. 
To reduce the excessively large search space, we did a pilot experiments on the optimizer hyperparameters, and found that setting learning rate to $1e^{-3}$ for dense and $1e^{-1}$ for sparse works the best for all models. We fixed the learning rate in the following runs.
We now describe model-specific search space:

\para{AFN+} The AFN hidden units and DNN hidden units are the same across all runs, followed the general MLP search space.
The number of logarithmic neurons ranges in $\{128, 256, 512, 1024\}$.

\para{AutoInt+} We created the search space based on the best configurations reported in the paper \cite{song2019autoint}, with a larger value being considered additionally per hyperparameter.
The number of attention layers ranged in $\{3, 4\}$, with attention dim ranged in $\{256, 512\}$.
The number of attention heads are in $\{4, 8\}$.
The DNN hidden units follow the general MLP search space.

\para{DCNv2} The number of cross layers ranged from 1 to 4. Rank searched in either full-rank or $512$.

\para{DLRM} The bottom MLP layer sizes and numbers was set to $[512, 256]$.

\para{FinalMLP} We followed the public benchmark setup \cite{bars}, by setting FeatureSelection (FS) to all float features for one stream, and searching over one of 8 selected sparse features for the other stream. FS MLP is set to $[800]$. Number of heads is fixed to $256$. 

\para{MaskNet} We tested both Parallel MaskNet and Serial MaskNet. For the Parallel variant, we consider the number of blocks in $\{1, 8, 16\}$ and the block dimension in $\{64, 128\}$. For the Serial variant, we consider the number of layers in $\{1, 4, 8\}$ with the layer size in $\{64, 256, 1024\}$. We fixed the reduction ratio to 1 for both variants. 

\para{xDeepInt} We considered Compressed Interaction Network (CIN) with the number of layers in $\{3, 4\}$ and the layer dimension in $\{16, 32, 64\}$.

\para{\wukong} The bottom MLP layer sizes and numbers was set to $[512, 256]$. 
$l$ ranged from $1$ to $4$; $n_F$ and $n_L$ are set to the same value, ranged in $\{8, 16\}$. 
$k$ is fixed to $24$.


\section{Model Complexity/Size on Public Datasets}
\label{sec:model-stats-public}
Please refer to Table \ref{tab:model-stats-public} for details.
\begin{table*}[]
\centering
\footnotesize\fontfamily{\robotofamily}\selectfont
\setlength{\tabcolsep}{1.6866em}
\begin{tabular}{lrrrrrr}
\toprule
         & \multicolumn{2}{c}{\textbf{Frappe}}                       & \multicolumn{2}{c}{\textbf{MicroVideo}}                   & \multicolumn{2}{c}{\textbf{MovieLens}}                    \\
         & \multicolumn{1}{l}{\#Params} & \multicolumn{1}{l}{MFLOP}  & \multicolumn{1}{l}{\#Params} & \multicolumn{1}{l}{MFLOP}  & \multicolumn{1}{l}{\#Params} & \multicolumn{1}{l}{MFLOP}  \\
\midrule


AFN+     & 13607961                     & 81.45                      & 3205771                      & 10.57                      & 11259899                     & 33.36                      \\
AutoInt+ & 306151                       & 15.98                      & 1989378                      & 10.74                      & 1293590                      & 3.00                       \\
DCNv2    & 7251073                      & 39.35                      & 3454209                      & 11.84                      & 1243101                      & 2.02                       \\
DLRM     & 1527073                      & 4.99                       & 2853761                      & 12.66                      & 1238421                      & 1.99                       \\
FinalMLP & 3115954                      & 10.66                      & 3020498                      & 9.25                       & 1040902                      & 0.81                       \\
MaskNet  & 21189593                     & 122.92                     & 2053249                      & 7.87                       & 2624571                      & 10.30                      \\
xDeepFM  & 66206                        & 0.01                       & 5147446                      & 5.91                       & 994439                       & 0.00                       \\
Wukong   & 7769589                      & 44.09                      & 2219409                      & 9.08                       & 17734369                     & 37.49                      \\

\addlinespace[2px]
\midrule


         & \multicolumn{2}{c}{\textbf{KuaiVideo}}                    & \multicolumn{2}{c}{\textbf{TaobaoAds}}                    & \multicolumn{2}{c}{\textbf{Criteo}}                       \\
         & \multicolumn{1}{l}{\#Params} & \multicolumn{1}{l}{MFLOP} & \multicolumn{1}{l}{\#Params} & \multicolumn{1}{l}{MFLOP} & \multicolumn{1}{l}{\#Params} & \multicolumn{1}{l}{MFLOP} \\

\midrule

AFN+     & 84013143                     & 10.95                      & 167888773                    & 13.45                      & 26436527675                  & 1826.50                    \\
AutoInt+ & 43937794                     & 79.18                      & 167395330                    & 28.41                      & 26142619137                  & 163.34                     \\
DCNv2    & 42636609                     & 9.10                       & 193472513                    & 166.19                     & 26179364097                  & 262.35                     \\
DLRM     & 41446513                     & 1.99                       & 42192033                     & 4.53                       & 26136307201                  & 4.62                       \\
FinalMLP & 580227666                    & 12.17                      & 85624114                     & 16.39                      & 26149500924	 & 84.95                       \\
MaskNet  & 41833034                     & 4.29                       & 42353201                     & 5.44                       & 26160209153                  & 147.40                     \\
xDeepFM  & 53912381                     & 6.81                       & 168867028                    & 2.98                       & 26300221171                  & 41.57                      \\
Wukong   & 44671649                     & 22.27                      & 175724173                    & 63.00                      & 26163636001  &	173.73                     \\

\bottomrule
\end{tabular}
\caption{
Model complexity and size on public datasets.
}
\label{tab:model-stats-public}
\end{table*}

\section{Model-Specific Scaling-up Configurations}
\label{appendix:scaleup}
Please refer to Table \ref{tab:scaleup-setup} for details.

\section{Analysis of High Order Interactions in Wukong}
The traditional factorization machine approach solves second order interaction problem by minimizing \cite{dlrm}: 
\begin{equation*}
\min \underset{i, j \in S}{\Sigma} r_{ij} - {X^1} {X^1}^T
\end{equation*}
where $r_{ij}\in R$ is the rating of the $i$-th product by the $j$-th user for $i$ = 1, ..., m and $j$ = 1, ..., n; X denotes the user and item representations (embeddings), and the superscript $1$ denotes the embedding contains $1$st order information. The dot product of these embedding vectors yields a meaningful prediction of the subsequent rating for 2nd order interactions. In \wukong, this meaningful interactions are then transformed to 2nd order interaction representations $X^2$ using MLP. In the 2nd layer FMB, with a residual and LCB connection, a dot product of  
(${X^1 + X^2}) (X^1 + X^2)^T$ yield both meaningful interaction from 1st order to 4th order. By analogy, a $l$-layer \wukong solves a problem by minimizing:
\begin{equation*}
\min \underset{i, j \in S}{\Sigma} (r_{ij} - \underset{k \in 1,2,...,2^{l-1}}{\Sigma}{X^k} {X^k}^T)
\end{equation*}
Thus, comparing to the traditional factorization approach, \wukong is able to solve the recommendation problem with a more sufficient interaction orders.

\section{Scaling Law on Training Data Volume}
\label{sec:qualityvsdatasetsize}
\begin{figure}[h]
    \centering
    \includegraphics[width=1\linewidth]{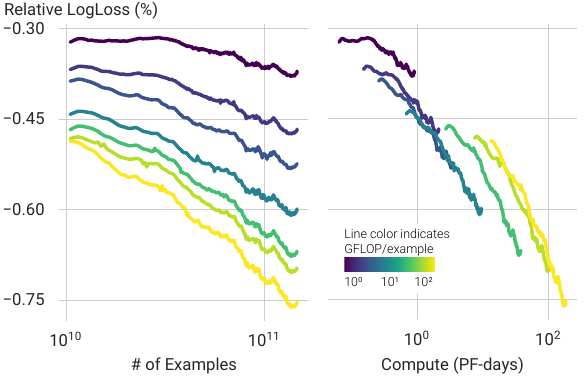}
    \caption{Wukong's model quality improvements versus training data volume and training compute.}
    \label{fig:ne-data}
\end{figure}

Fig.~\ref{fig:ne-data} provides a summary for \wukong's performance versus the dataset size on which it is trained on (one pass). Similar to what has been observed on LLMs, we found that large models are more data-efficient, meaning that they require fewer samples to achieve the same quality improvement. In addition, we found that all \wukong models have consistently improved their model quality up to the end of 146B data, while larger models have a
steeper trend in the model quality improvement. We also noticed that one of the limitations of our study is the dataset size is still far less for the large model to converge, which will be one of the areas for further study.


\section{Comparing with Transformer-based Approaches}
We highlight differences and provide intuitions on why \wukong scales better than Transformer-based approaches like AutoInt+~\cite{song2019autoint}. While the structure of \wukong resembles that of Transformer's, we note the following architectural difference: first, the projection used in \wukong is MLP (bit-wise) in both FMB and each layer instead of FFN (embedding/position-wise) in transformer; second, \wukong is configured as a pyramid shape versus the uniform shape used in transformers.

We hypothesize that the difference in projection plays an important role in quality delivery. These MLPs operate over the flattened input embeddings, essentially providing each feature with a different projection matrix. Our intuition is that this helps the model learn from heterogeneous input features, contrasting to a single embedding space used in LLMs. Similar intuition was discussed in~\cite{gui2023hiformer}.

Efficiency-wise, we argue that the pyramid shape configuration allows \wukong to exclude unnecessary computations by contracting the number of embeddings used in each layer. 

To verify these hypothesis, we conduct the following experiments by applying Wukong’s unique components to Autoint+, which conclude (1) using bit-wise MLP instead of FFN for V-projection improves LogLoss by 0.34\%; (2) adding bit-wise MLPs after self-attentions improves LogLoss by 0.65\%; (3) combining both, along with a pyramid layer shape (by using LCB on the first layer’s output) achieves 0.57\% quality improvement. Compared to the scaled up Autoint+, \wukong achieves 0.08\% quality improvement, while saves 90\% FLOPs.
We summarize the results in Table \ref{tab:comparison-autoint}

\begin{table}[h]
\footnotesize\fontfamily{\robotofamily}\selectfont
\setlength{\tabcolsep}{.88666em}
\begin{tabular}{lrr}
\toprule
     \textbf{Changes} & \textbf{Relative}     & \textbf{GFLOP}        \\
     vs. Autoint+ & \textbf{LogLoss} (\%) &  \textbf{/example} \\
\midrule

\addlinespace[2px]

Vanilla Autoint+ & 0 & 8 \\

\addlinespace[2px]
\hdashline
\addlinespace[4px]

V=FFN() $\to$ V=MLP() & -0.34 & 13 \\

\addlinespace[2px]
\hdashline
\addlinespace[4px]

Scaled-up Autoint+ & -0.49 & 50 \\

\addlinespace[2px]
\hdashline
\addlinespace[4px]

V=FFN() $\to$ V=MLP() & -0.57 & 5 \\
Layer FFN $\to$ Layer MLP \\
Pyramid layer shape \\

\addlinespace[2px]
\hdashline
\addlinespace[4px]
Layer FFN $\to$ Layer MLP & -0.65 & 36 \\  

\addlinespace[2px]

\bottomrule
\end{tabular}
\caption{
Replacing/Adding Wukong's unique components to vanilla Autoint+ improves the model quality.
}
\label{tab:comparison-autoint}
\end{table}

\begin{table*}[hbt]
\centering
\footnotesize\fontfamily{\robotofamily}\selectfont
\setlength{\tabcolsep}{0.39666em}
\begin{tabular}{lrrrr}
\toprule
  \textbf{Hyperparameters}   & \textbf{\gflop}   & \textbf{\nparams}  & \textbf{\rll}    & \textbf{\rll}    \\
        &  & (B) & (Task1) & (Task2)   \\
\midrule

\addlinespace[2px]
\multicolumn{3}{l}{\textit{AFN+}} \\
\addlinespace[2px]

DNN=4x2048, afn=4x2048, nlog=1024                                                                                                               & 4.41    & 628.22   & 0.11  & 0.05         \\
DNN=4x4096, afn=4x2048, nlog=1024                                                                                                               & 7.65    & 628.74   & 0.12  & 0.06         \\
DNN=4x4096, afn=4x4096, nlog=2048                                                                                                               & 13.08   & 629.46   & 0.21  & 0.14         \\
DNN=4x8192, afn=4x8192, nlog=4096                                                                                                               & 43.4    & 633.95   & 0.12  & 0.06         \\

\addlinespace[2px]
\hdashline
\addlinespace[4px]
\multicolumn{3}{l}{\textit{AutoInt+}} \\
\addlinespace[2px]

Attention=3x256, nhead=4, DNN=2x256                                                                                                              & 7.72    & 627.73   & 0.39 & 0.24          \\
Attention=3x512, nhead=4, DNN=2x256                                                                                                              & 18.58   & 627.77   & 0.15 & 0.05          \\
Attention=3x512, nhead=8, DNN=3x8192                                                                                                             & 42.53   & 631.49  & -0.09 & -0.16         \\
Attention=3x512, nhead=16, DNN=3x10240                                                                                                             & 49.58	& 632.59  & -0.1 & -0.2	          \\
Attention=3x512, nhead=16, DNN=3x16384                                                                                                             & 68.83	& 635.57  & 0.13 (LossX) & 0.01 (LossX)	         \\

\addlinespace[2px]
\hdashline
\addlinespace[4px]
\multicolumn{3}{l}{\textit{DCN}} \\
\addlinespace[2px]

l=2, rank=512, MLP=4x2048                                                                                                                          & 3       & 628.11 & -0.27 & -0.27          \\
l=2, rank=512, MLP=4x4096                                                                                                                          & 4.67    & 628.37 & -0.29 & -0.32         \\
l=2, rank=512, MLP=4x16384                                                                                                                         & 17.85   & 630.42  & -0.38 & -0.41 \\
l=2, rank=512, MLP=4x32768                                                                                                                         & 43.88   & 634.46  & -0.43 & -0.45          \\
l=2, rank=512, MLP=4x51200                                                                                                                & 84.71    & 640.79 & (LossX)  & (LossX)           \\

\addlinespace[2px]
\hdashline
\addlinespace[4px]
\multicolumn{3}{l}{\textit{DLRM}} \\
\addlinespace[2px]

TopMLP=2x512                                                                                                                                           & 1.37 & 627.78 &  \em{(Baseline)}     &  \em{(Baseline)}              \\
TopMLP=4x512                                                                                                                                           & 1.37    & 627.78  & -0.11 & -0.08          \\
TopMLP=4x2048                                                                                                                                          & 3.85    & 628.17  & -0.23 & -0.21         \\
TopMLP=4x4096                                                                                                                                          & 7.29    & 628.7   & -0.28 & -0.27          \\
TopMLP=4x8192                                                                                                                                          & 14.61   & 629.84  & -0.32 & -0.31         \\
TopMLP=4x16384                                                                                                                                         & 31      & 632.39  & -0.37 & -0.35          \\
TopMLP=4x32768                                                                                                                                         & 71.23   & 638.62  & -0.36 & -0.34          \\

\addlinespace[2px]
\hdashline
\addlinespace[4px]
\multicolumn{3}{l}{\textit{FinalMLP}} \\
\addlinespace[2px]

MLP1=4x4096, MLP2=2x1024, output\_dim=64,                                                      & 3.93    & 628.25  & -0.11 & -0.16          \\
\hspace{1ex}no\_fs \\
MLP1=4x4096, MLP2=2x1024, output\_dim=64,   & 8.17    & 628.91  & -0.23 & -0.27         \\
\hspace{1ex}fs1={[}0,57600{]}, fs2={[}57600,115200{]}, fs\_MLP=1x2048 \\
MLP1=4x8192, MLP2=2x2048, output\_dim=64,   & 16.9    & 630.27  & -0.34  & -0.36         \\
\hspace{1ex}fs1={[}0,57600{]}, fs2={[}57600,115200{]}, fs\_MLP=1x4096 \\
MLP1=8x8192, MLP2=4x2048, output\_dim=64,   & 18.77   & 630.56  & -0.37  & -0.38         \\
\hspace{1ex}fs1={[}0,57600{]}, fs2={[}57600,115200{]}, fs\_MLP=2x4096, \\
MLP1=4x16384, MLP2=2x4096, output\_dim=64,  & 36.26   & 633.27 & -0.34 & -0.34          \\
\hspace{1ex}fs1={[}0,57600{]}, fs2={[}57600,115200{]}, fs\_MLP=1x8192 \\
MLP1=4x24576, MLP2=2x6144, output\_dim=64, & 58.12   & 636.67  & -0.37 & -0.38          \\
\hspace{1ex}fs1={[}0,57600{]}, fs2={[}57600,115200{]}, fs\_MLP=1x12288 \\

\addlinespace[2px]
\hdashline
\addlinespace[4px]
\multicolumn{3}{l}{\textit{MaskNet}} \\
\addlinespace[2px]

MLP=1x512, nblock=1, dim=128, reduction=0.01                                                                                                    & 1.76    & 627.92  & -0.09 & -0.12          \\
MLP=1x512, nblock=4, dim=128, reduction=0.01                                                                                                    & 6.8     & 628.7   & -0.22  & -0.25        \\
MLP=3x2048, nblock=4, dim=128, reduction=0.01                                                                                                   & 6.88    & 628.71   & -0.28 & -0.3         \\
MLP=3x2048, nblock=4, dim=128, reduction=0.05                                                                                                   & 32.36   & 632.67 & -0.37 & -0.37          \\
MLP=3x2048, nblock=4, dim=128, reduction=0.1                                                                                                    & 64.21   & 637.61 & -0.4  & -0.4      \\

\addlinespace[2px]
\hdashline
\addlinespace[4px]
\multicolumn{3}{l}{\textit{Wukong}} \\
\addlinespace[2px]

l=2, nL=8, nF=8, k=24, MLP=3x2048                                                                                                               & 0.53    & 627.74  & -0.35  & -0.32        \\
l=4, nL=32, nF=32, k=24, MLP=3x2048                                                                                                             & 1.25    & 627.82  & -0.45  & -0.43        \\
l=8, nL=32, nF=32, k=24, MLP=3x2048                                                                                                             & 2.12    & 627.95  & -0.53  & -0.49        \\
l=8, nL=48, nF=48, k=48, MLP=3x4096                                                                                                             & 5.6     & 628.46 & -0.6  & -0.6           \\
l=8, nL=96, nF=96, k=96, MLP=3x8192                                                                                                             & 22.23   & 630.96  & -0.67 & -0.66          \\
l=8, nL=96, nF=96, k=96, MLP=3x16384                                                                                                            & 61      & 636.99  & -0.7 & -0.69          \\
l=8, nL=192, nF=192, k=192, MLP=3x16384                                                                                                         & 108     & 644    & -0.76 & -0.76          \\

\bottomrule

\end{tabular}
\caption{Detailed hyperparameters, compute complexity, model quality and model size for each run evaluated in Sec. \ref{sec:eval-prod}. LossX means loss exploded during training.}
\label{tab:scaleup-setup}
\end{table*}

\end{document}